\pdfoutput=1
\documentclass[letterpaper]{article}
\PassOptionsToPackage{table}{xcolor}
\usepackage[preprint]{aaai2027}
\usepackage[hyphens]{url}
\usepackage{graphicx}
\usepackage{natbib}
\usepackage{caption}
\usepackage{multirow}

\usepackage{amssymb}
\usepackage{amsmath}
\usepackage{enumitem}
\usepackage{xspace}
\usepackage{array}
\usepackage{booktabs}
\usepackage{microtype}
\setlist[itemize]{label=\textbullet, leftmargin=1.5em}

\definecolor{TabBG}{RGB}{235,245,255}

\newcommand{\system}{RoadState\xspace}
\newcommand{\model}{RoadState-VLM\xspace}

\title{\system{}: A Domain-Prior-Enhanced Vision-Language System for Industrial-Grade Lane-Level Map Updating}

\author{
    Deguo Xia\textsuperscript{\rm 1,\rm 2}\equalcontrib,
    Zihan Li\textsuperscript{\rm 3}\equalcontrib,
    Jiaqi Wang\textsuperscript{\rm 2}\equalcontrib,
    Dong Xie\textsuperscript{\rm 2},
    Xiyan Liu\textsuperscript{\rm 2},
    Chen Zhu\textsuperscript{\rm 2},
    Liwei Qian\textsuperscript{\rm 2},
    Xin Pei\textsuperscript{\rm 2},
    Shuanglong Li\textsuperscript{\rm 2},
    Hanmeng Liu\textsuperscript{\rm 2},
    Jizhou Huang\textsuperscript{\rm 2}\corresponding,
    Mengmeng Yang\textsuperscript{\rm 1}\corresponding,
    Diange Yang\textsuperscript{\rm 1}
}
\affiliations{
    \textsuperscript{\rm 1}Tsinghua University, Beijing, China\\
    \textsuperscript{\rm 2}Baidu Inc., Beijing, China\\
    \textsuperscript{\rm 3}University of Macau, Macau, China\\
    xdg23@mails.tsinghua.edu.cn, \{xiedong04, huangjizhou01\}@baidu.com, yangmm\_qh@tsinghua.edu.cn
}

\begin{document}
\maketitle

\begin{abstract}

Lane-level maps provide foundational structured road information for lane-level navigation, intelligent driving, and intelligent transportation systems. Understanding road-scene semantics from structural and contextual cues is essential for reliable lane-level map updating. However, this task remains challenging in production because road scenes involve occlusions, ambiguous lane markings, temporary obstacles, and viewpoint changes, while downstream map-maintenance workflows require temporally consistent, physically valid, and auditable outputs rather than free-form visual descriptions.
We present \system{}, a production-deployed vision-language system that transforms sequential road-scene observations into structured lane-level semantics for Baidu Maps' map-updating pipeline. Its core model, \model{}, formulates road-scene understanding as autoregressive state estimation, predicting the current structured road state from the previous observation and symbolic state together with the current observation. To improve semantic consistency and reasoning fidelity across interdependent road-state dimensions, we introduce Geometric Chain-of-Thought (Geo-CoT), which uses a directed acyclic graph of semantic dependencies as a domain prior for structured reasoning.
Extensive experiments on a real-world dataset show that \model{} achieves 94.33\% mean accuracy across six road-state dimensions, outperforming the strongest fine-tuned VLM baseline.
Ablation studies further demonstrate the complementary effectiveness of Geo-CoT and temporal transition context.
For production deployment, a cascaded routing pipeline directs routine cases to lightweight models while reserving complex cases for \model{}. Since its deployment in March 2025, \system{} has supported lane-level map updating across more than 360 cities and reduced overall production costs by 30\%. 
Together with human-in-the-loop auditing, the system maintains robust performance under stress-test conditions. 
These results demonstrate how domain-prior-enhanced reasoning, structured state interfaces, selective routing, and human quality control can bridge the gap between VLM-based road-scene understanding and large-scale lane-level map updating.

\end{abstract}

\begin{figure}[t]
\centering
\includegraphics[width=\linewidth, height=0.58\linewidth, clip]{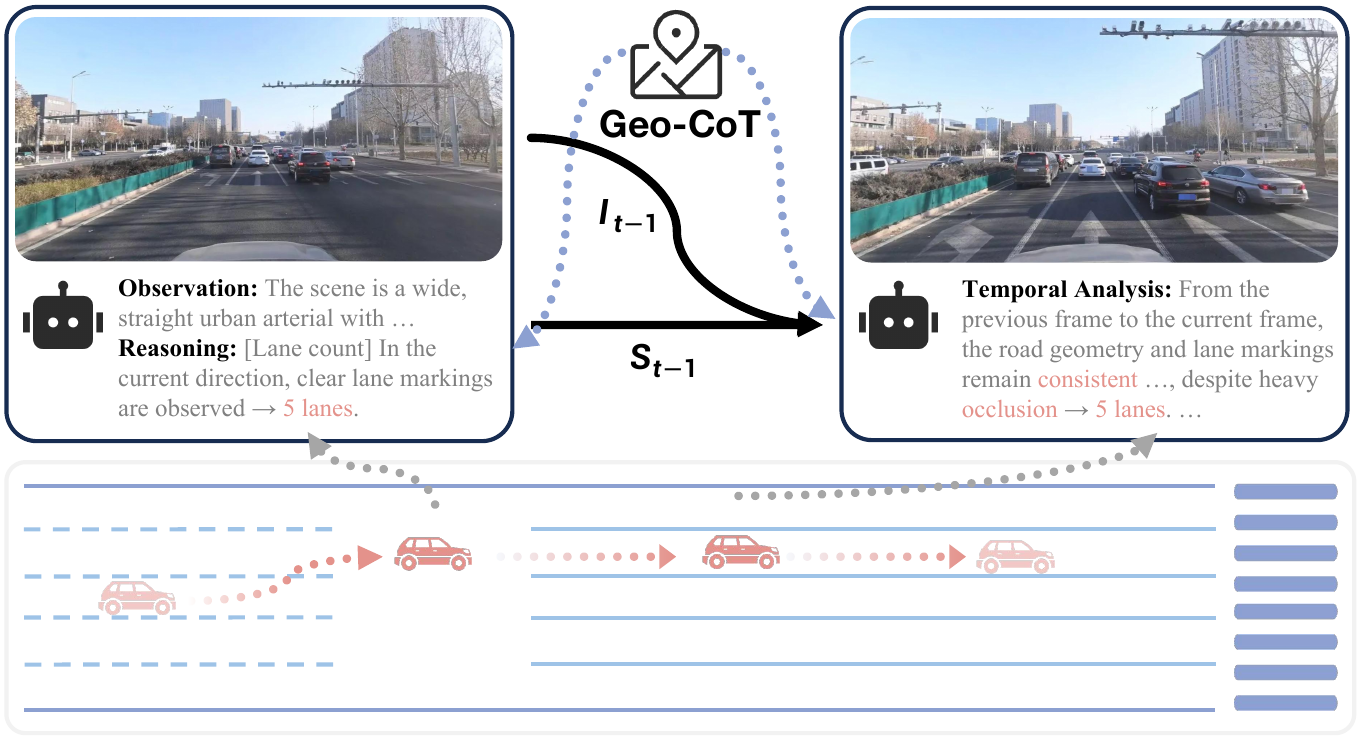}
\caption{An example of structured road information generated from a crowdsourced front-view image sequence.}
\label{fig:ad}
\end{figure}

\section{Introduction}
\label{sec:intro}

Lane-level maps provide structured road information for lane-level navigation, intelligent driving, and intelligent transportation systems \cite{elghazaly2023high,xia2024dumapnet,xia2025ldmapnetu,xia2026mapagent}. Industrial systems have used street-view imagery and learned vector generation to detect map changes and generate or update city-scale lane maps \cite{xia2026end,tang2025priorfusion}. Public driving datasets provide supervision for 2D scene parsing and multimodal 3D detection, tracking, and forecasting \cite{Cityscapes8,yu2020bdd100k,nuscenes4,chang2019argoverse,wilson2023argoverse}, while online HD-mapping methods reconstruct vectorized map elements from onboard sensors \cite{li2022hdmapnet,liu2023vectormapnet,liao2022maptr,liao2025maptrv2}. These advances establish low-level perception and metric geometry, whereas map updating also requires relational semantics: how lanes are organized, where the observing vehicle is located, and what constrains movement between adjacent lanes \cite{liu2026roadscenebench}. In Baidu Maps' workflow, these structured states support observation selection and interpretation of candidate map changes (Figure~\ref{fig:ad}). Crowdsourced front-view sequences supply frequent observations, but complex scenes still require substantial manual interpretation.

Contrastive pretraining, multimodal few-shot learning, and visual instruction tuning have substantially expanded general visual-language representation and generation \cite{radford2021learning,li2022blip,alayrac2022flamingo,dai2023instructblip}. Driving-oriented VLMs further support action explanation, graph-based scene reasoning, and closed-loop control \cite{drivelm,xu2025drivegpt4}, while VideoChat and Video-LLaVA extend multimodal interaction to video \cite{li2023videochat,lin2024videollava}. 
Driving QA and fine-grained VLM benchmarks evaluate detailed scene reasoning \cite{qian2024nuscenes,li2025fine}, and temporal hallucinations remain measurable in video MLLMs \cite{li2025vidhalluc}. 
Applying these capabilities to map updating raises four connected requirements. First, frame-wise inference lacks an explicit road-state estimate across observations. Under occlusion or ambiguous markings, unsupported fluctuations in lane count or ego-lane index can produce false change candidates \cite{liu2026roadscenebench}; temporal evidence can resolve some of these ambiguities \cite{zhang2021vil}. Second, supervision on final labels leaves cross-field dependencies implicit: the ego-lane index must be valid for the lane count, and lane-change feasibility must combine structural, vehicle, and obstacle constraints. Third, reviewers need concise, image-grounded reasons to verify and correct individual fields under a time budget. Finally, feasibility at daily-update volume depends on the combined per-frame cost of model serving and human review. 

To meet these production requirements, we present \system{}, a vision-language system deployed in Baidu Maps' map-updating workflow to generate structured lane-level semantics.
Its core model, \model{}, formulates road-scene understanding as autoregressive state estimation: it predicts the current state from the previous image, previous symbolic state, and current image. The symbolic state provides a compact temporal prior, while the image pair supplies evidence for retaining or revising it. Geometric Chain-of-Thought (Geo-CoT) distillation encodes a domain-defined dependency graph from visual observations through lane topology and ego localization to maneuver constraints, yielding structured supervision and reviewable intermediate reasoning.
For production deployment, \system{} uses cascaded inference to route routine observations to lightweight models and complex cases to \model{}.
Reviewers inspect flagged images and fix individual fields.

On our real-world road-scene dataset, \model{} achieves 94.33\% mean accuracy across six state dimensions, outperforming the strongest fine-tuned VLM baseline by 2.23 percentage points. Ablation studies show complementary gains from Geo-CoT and temporal transition context. Since its deployment in March 2025, \system{} has supported lane-level map updating in more than 360 cities, increased annotation throughput by 43.8\%, and reduced overall production costs by 30\%, demonstrating both model effectiveness and production-scale utility.

The main contributions of this work are:
\begin{itemize}
\item A deployed lane-level semantic-analysis system that connects cascaded model inference, a shared state interface, and human review to map updating.
\item An autoregressive road-state model with DAG-guided Geo-CoT supervision, combining explicit semantic dependencies with historical state feedback.
\item Technical evaluation and production measurements that characterize the effects of the model design and its impact on map-annotation efficiency.
\end{itemize}

\section{\system{} System}
\label{sec:system}

\begin{figure*}[t]
  \centering
  \includegraphics[width=\textwidth]{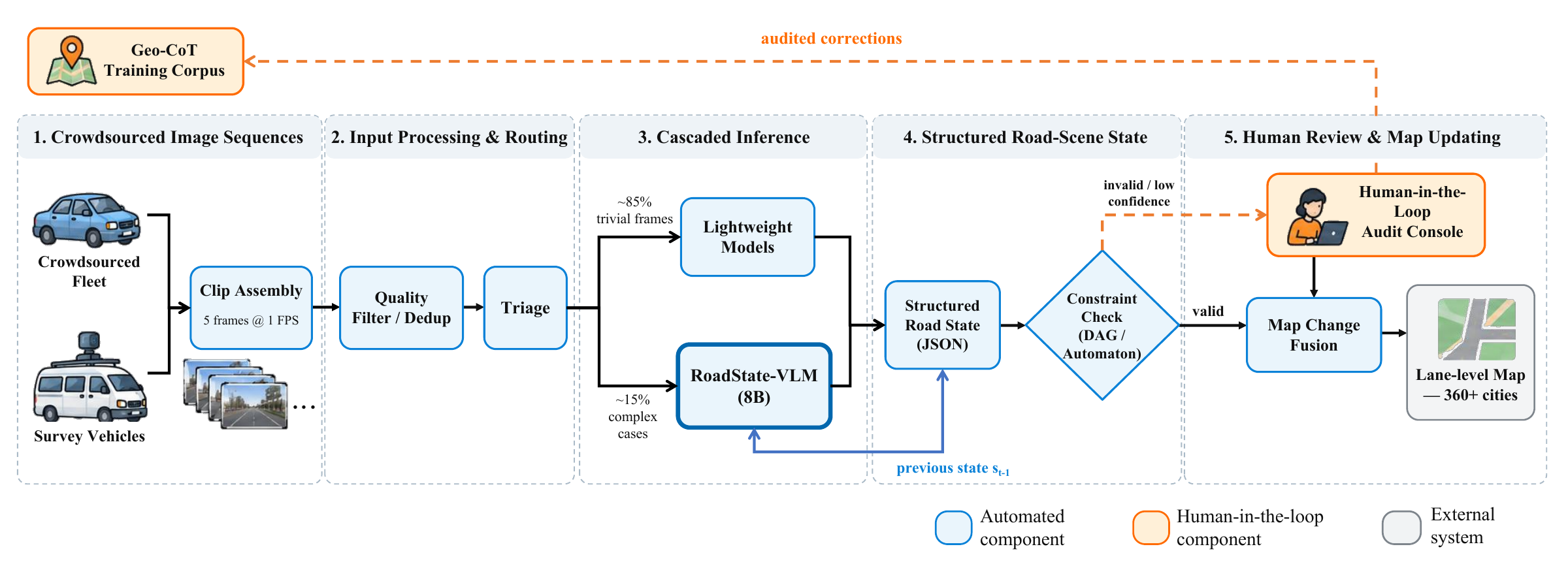}
  \caption{The \system{} production system. Crowd-sourced and survey imagery is assembled into clips, triaged by complexity, and routed to either lightweight models ($\sim$85\% of frames) or \model{} ($\sim$15\%). Both branches emit the same structured road state, which must pass a deterministic constraint check before entering map change fusion; violations and low-confidence outputs are escalated to human auditors, whose corrections are recycled as Geo-CoT training data. Section~\ref{sec:framework} details the \model{} component.}
  \label{fig:system_overview}
\end{figure*}

\subsection{Application Context and Production Requirements}
\label{sec:application_context}
Lane-level maps at Baidu Maps are maintained by continuously detecting road changes from monocular front-view imagery. Survey fleets equipped with LiDAR-camera suites yield near-gold-standard geometry but are too costly to revisit every road segment frequently, so crowd-sourced and light survey imagery carries most of the change signal. Crucially, this pipeline is an \emph{offline batch} process rather than onboard perception: a frame may be processed seconds after capture, which relaxes latency budgets but makes auditability and per-frame cost the binding constraints.

Our prior work, RoadSceneBench~\cite{liu2026roadscenebench}, established the mid-level semantic representation for this task across 2{,}341 five-frame front-view clips (11{,}705 images from 20 Chinese cities), covering interdependent attributes such as lane count, ego-lane index, and lane-change feasibility. However, RoadSceneBench was designed for offline model evaluation. When piloting frame-wise perception models inside Baidu Maps' daily update pipeline, we encountered four operational bottlenecks across downstream change detection, manual quality control, and serving infrastructure, which directly defined our system requirements:

\begin{itemize}[leftmargin=*, itemsep=2pt, topsep=3pt]
    \item \textbf{(R1) Structural validity.} Predictions must satisfy lane-topology rules; for instance, predicting ego-lane 4 on a 3-lane road corrupts downstream observation filtering even if each attribute looks plausible in isolation.
    \item \textbf{(R2) Temporal stability.} Consecutive observations of the same road segment must not flicker in lane count or ego-lane index under brief occlusion, since downstream change detection treats state differences as candidate real-world road changes.
    \item \textbf{(R3) Auditability.} Human auditors verify flagged predictions under a strict time budget; providing final labels without concise visual reasons forces them to re-inspect the entire scene from scratch.
    \item \textbf{(R4) Throughput and cost.} At daily city-scale volume, applying a heavy VLM or manual review to every frame is prohibitively expensive, so combined serving and review cost per frame must remain low.
\end{itemize}

These four requirements directly shaped the design of \system{}: autoregressive state transition addresses R2, DAG-guided Geo-CoT supervision addresses R1 and R3, and cascaded routing with human-in-the-loop review addresses R4.

\subsection{\system{} System Architecture}
\label{sec:system_architecture}

Figure~\ref{fig:system_overview} shows the deployed system; the \model{} component inside it is the subject of Section~\ref{sec:framework}. Incoming imagery is first assembled into short clips along each trajectory, following the five-frame organization of RoadSceneBench, and passed through automated quality filtering and deduplication. A triage step then routes each clip: roughly $85\%$ of frames depict unambiguous geometry and are handled by lightweight classifiers, while the remaining $15\%$---occlusions, ambiguous markings, ramps, temporary obstacles---are escalated to \model{}, which additionally receives the previous symbolic state as a temporal anchor. Both branches emit the same structured road state, so downstream stages are agnostic to which model produced it.

Every state then passes a deterministic constraint check derived from the same dependency graph used to supervise Geo-CoT. States that satisfy all constraints become candidate map changes; violations and low-confidence outputs are escalated to an audit console, where the structured state and its intermediate reasoning are presented for correction. Accepted changes enter map change fusion and the lane-level map update workflow. Audited corrections are recycled as Geo-CoT training data, closing a second, slower loop that improves the model as the pipeline runs.

\section{\model{} Framework}
\label{sec:framework}
\begin{figure*}[!t]
  \centering
  \includegraphics[width=0.9\textwidth]{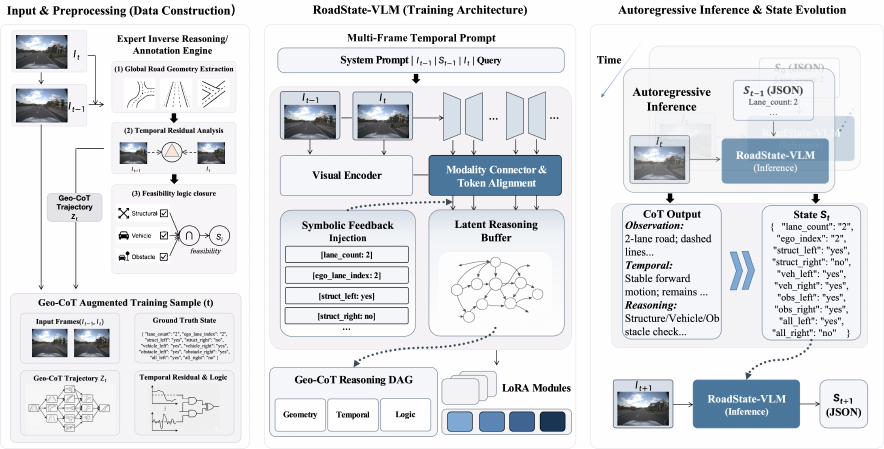}
  \caption{Overview of the \model{} framework. The pipeline consists of three main stages covering input and preprocessing for Geo-CoT data construction, the RoadState-VLM training architecture with symbolic feedback injection and a latent reasoning buffer, and autoregressive inference with state evolution for temporally consistent road state estimation.}
  \label{fig:framework_overview}
\end{figure*}

As shown in Figure~\ref{fig:framework_overview}, we cast road-scene understanding as symbol-prioritized state estimation: the VLM predicts a structured road state from the current frame, the previous frame, and the previous symbolic state. Historical feedback acts as a soft anchor, while geometric priors constrain the output space for temporal coherence and physical validity.

\subsection{Problem Setup}
\label{sec:Problem_Setup}
Instead of mapping each frame $I_t$ independently to a textual description, we estimate a sequence of structured road states $\mathbf{S} = \{s_1, s_2, \dots, s_T\}$ from driving images $\mathcal{I} = \{I_1, I_2, \dots, I_T\}$. We use a first-order Markovian transition to capture local temporal dynamics efficiently:
\begin{equation}
    \hat{s}_t = \arg\max_{s \in \mathcal{S}_{state}} P(s \mid I_t, I_{t-1}, s_{t-1}; \theta),
\end{equation}
where $\hat{s}_t$ is the predicted structured state, $\mathcal{S}_{state}$ is the discrete state space, and $s_{t-1}$ serves as the symbolic prior. This formulation transforms the VLM into an autoregressive state operator, allowing the model to utilize the previous state $s_{t-1}$ as a ``soft anchor'' to ensure temporal smoothness and logical continuity.

\textbf{Definition of State Space.}
The state $s_t$ contains three core dimensions:
\begin{itemize}
    \item \textbf{Lane Topology Count} ($C \in \{1, \dots, N\}$): Represents the number of motor-vehicle lanes in the current travel direction.
    \item \textbf{Ego-Lane Index} ($I \in \{1, \dots, C\}$): Identifies the lane occupied by the ego vehicle, indexed from left to right (e.g., $1$ is the leftmost lane).
    \item \textbf{Multi-Dimensional Feasibility} ($\mathcal{F}$): Decomposes lane-change feasibility into three layers for each candidate direction:
    \begin{itemize}
        \item \textit{Structural} ($\mathcal{F}_{str}$): Physical constraints such as solid versus dashed lane markings and road boundaries.
        \item \textit{Vehicle} ($\mathcal{F}_{veh}$): Dynamic constraints imposed by surrounding traffic flow and safety gaps.
        \item \textit{Obstacle} ($\mathcal{F}_{obs}$): Clearance constraints from static or temporary objects (e.g., traffic cones or pedestrians).
    \end{itemize}
    The global feasibility is defined as the logical conjunction:
    \begin{equation}
        \mathcal{F}_{all} = \mathcal{F}_{str} \wedge \mathcal{F}_{veh} \wedge \mathcal{F}_{obs}.
    \end{equation}
\end{itemize}

\textbf{Data Preparation.} We curate RoadSceneState by extending RoadSceneBench with 2,341 five-frame temporal clips (11,705 images, sampled at 1 Hz). Geo-CoT trajectories provide transition-level supervision from visual observations to structured states.

\subsection{Geo-CoT Distillation}

A pivotal component of our framework is the construction of Geometric Chain-of-Thought (Geo-CoT) trajectories. Although the structured state $s_t$ provides a compact representation for downstream processing, the state tokens alone do not expose the visual evidence and intermediate geometric decisions that support a transition. We therefore adopt an expert-informed inverse-distillation strategy using the multimodal teacher model \textbf{GPT-5.2}. Given the consecutive observations $\{I_{t-1}, I_t\}$, the historical state $s_{t-1}$, and the target ground-truth state $s_t^{\mathrm{GT}}$, the teacher reconstructs an evidence-grounded reasoning trajectory that explains how the target transition follows from the observed scene. The ground-truth state is used only to construct supervision during data preparation and is unavailable to the student at inference time. Thus, Geo-CoT serves as structured rationale supervision rather than ordinary free-form prompting.

The reasoning template follows the four-node dependency DAG ``Observation $\rightarrow$ Topology $\rightarrow$ Localization $\rightarrow$ Constraints'', operationalized in three stages. First, \textit{geometric observation and topology grounding} identifies road boundaries, lane markings, and traffic participants to estimate the number of lanes in the current travel direction and the valid topological layout. Second, \textit{temporal localization and state maintenance} compares $I_{t-1}$ and $I_t$ to determine the ego-lane index and detect transition cues, such as lateral motion relative to lane markings, boundary crossing, lane merging, or lane splitting. The historical state $s_{t-1}$ acts as a soft prior in this stage, but clear current-frame evidence may override it. Third, \textit{multi-dimensional constraint closure} separately evaluates structural accessibility $\mathcal{F}_{str}$, vehicle clearance $\mathcal{F}_{veh}$, and obstacle clearance $\mathcal{F}_{obs}$. For each candidate lane-change direction, global feasibility is determined by the logical conjunction:
\begin{equation}
    \mathcal{F}_{all} = \mathcal{F}_{str} \wedge \mathcal{F}_{veh} \wedge \mathcal{F}_{obs}.
\end{equation}

\subsection{\model{} Architecture}
\label{sec:architecture}
\model{} implements the above transition operator with a Qwen3-VL-8B-Thinking backbone and parameter-efficient adaptation. The system maps unstructured visual observations to structured states through Geo-CoT reasoning and symbolic feedback.

\paragraph{Reasoning Backbone: Thinking-Integrated VLM} 
We use Qwen3-VL-8B-Thinking as the reasoning engine because it supports visual grounding, multimodal alignment, and internal CoT generation. Raw $4K$ imagery is downsampled to at most $1,568,000$ pixels and projected into the LLM embedding space through the modality connector $\mathcal{C}$.

\paragraph{Input Representation: The Symbolic Feedback Loop.}
The temporal prompt interleaves historical context with current visual observations:
\begin{equation}
    X_{input} = [\text{System\_Prompt}, I_{t-1}, s_{t-1}, I_t, \text{Query}],
\end{equation}
where $I_{t-1}$ and $I_t$ are consecutive frames and $s_{t-1}$ is the serialized previous road state.

Re-injecting $s_{t-1}$ creates a Symbolic Feedback Loop: the model uses prior topology and ego localization as a soft anchor, then focuses on differential cues such as marking changes or lateral shifts. This improves semantic persistence and reduces prediction flickering.

\paragraph{Reasoning-Guided State Update.}

The inference process within the LLM backbone follows a hierarchical dual-stage execution logic, designed to transform transient visual observations into logically-grounded state evolutions:

\begin{itemize}
    \item \textbf{Latent Reasoning Stage (Geo-CoT):} The model generates a Geo-CoT trajectory $\mathcal{Z}$ that reconciles visual residuals between $I_{t-1}$ and $I_t$ with constraints inherited from $s_{t-1}$, including the closure over Structural, Vehicle, and Obstacle feasibility.

    \item \textbf{Symbolic Conclusion Stage (JSON):} The model collapses the reasoning trajectory into a compact JSON state $s_t$, yielding an output that is both human-interpretable for auditing and machine-actionable for downstream modules.

\end{itemize}

\paragraph{Autoregressive Inference and Closed-Loop Evolution}

In practical deployment, \model{} operates in an autoregressive manner, transforming the continuity of perception into a recursive evolution of discrete states. 

The initial state $s_0$ is obtained by applying structured perception directly to the first frame, without relying on any historical prior. At each subsequent time step $t > 0$, the model uses its own predicted state $\hat{s}_{t-1}$ from the previous step as the current input prior. This autoregressive formulation is formalized as:
\begin{equation}
\hat{s}_t = \mathcal{F}_{vlm}(I_t, I_{t-1}, \hat{s}_{t-1}; \theta_{LoRA}).
\end{equation}

We reset propagation every five frames to bound local errors and prevent long-range drift. The previous state remains a soft anchor: when visual evidence in $I_t$ conflicts with history, the model can override the prior and usually recovers within one frame, or within 1--2 frames in ambiguous cases.

\subsection{Learning and Optimization Objectives}
\label{sec:optimization}

Training combines sequence supervision, structural constraints, and temporal alignment to turn the pre-trained backbone into a physics-aware state machine.

\paragraph{Reasoning-Aware Sequence Modeling.}
We formalize the perception task as a likelihood estimation of a constrained reasoning trajectory $\mathcal{T}$, which couples the Geometric Chain-of-Thought (Geo-CoT) $\mathcal{Z}$ with the final structured state $s_t$. The core objective is to minimize the negative log-likelihood (NLL) of the sequence:
\begin{equation}
    \mathcal{L}_{\mathrm{sft}} = - \mathbb{E}_{(X, \mathcal{T}) \sim \mathcal{D}} \sum_{j=1}^{M} \log P( \tau_j \mid X_{input}, \tau_{<j}; \theta \oplus \Delta\theta ),
\end{equation}
where $\tau_j$ denotes each token in the trajectory. Supervising the full trajectory encourages evidence-grounded reasoning before state prediction.

\paragraph{Constraint-Aware State Prediction.}

We incorporate axiomatic priors $\mathcal{P}_{logic}$ from the \textit{Road Structural Automaton}, such as lane-count conservation and ego-position boundaries, so the model learns to project visual features onto the valid state space $\mathcal{S}_{valid}$:
\begin{equation}
    \hat{s}_t = \arg\max_{s \in \mathcal{S}_{valid}} P(s \mid I_t, I_{t-1}, s_{t-1}, \mathcal{P}_{logic}),
\end{equation}
where $\mathcal{P}_{logic}$ represents structural axioms that enforce logical closure.

\paragraph{Temporal Consistency and Motion Grounding.}
We align a normalized symbolic mutation indicator with the visual residual between consecutive frames:

\begin{equation}
    \mathcal{L}_{temp} = \left| \Delta_s(s_t, s_{t-1}) - \widetilde{d}_{\Phi}(I_t, I_{t-1}) \right|,
\end{equation}

where $\Delta_s(s_t, s_{t-1}) \in [0,1]$ is a scalar indicator of semantic mutation (0 when the structured state is maintained and 1 when a topology, ego-lane, or lane-boundary transition occurs). We normalize the cosine distance by defining $\widetilde{d}_{\Phi}(I_t, I_{t-1}) = \tfrac{1}{2}\,\text{dist}_{\Phi}(\Phi(I_t),\Phi(I_{t-1})) \in [0,1]$, where $\Phi(\cdot)$ is the vision encoder and $\text{dist}_{\Phi}$ is the cosine distance. This makes the two terms comparable and penalizes symbolic flickering when the visual residual is small.

Because the structured state is discrete, we do not backpropagate through a discrete $\arg\max$. Instead, $\Delta_s$ is serialized as an auxiliary transition target in the Geo-CoT supervision and optimized through the corresponding token-level SFT term. Thus, $\lambda \mathcal{L}_{temp}$ denotes the weighted auxiliary alignment loss rather than a gradient through discrete state tokens. The full objective is

\begin{equation}
    \mathcal{L}_{total} = \mathcal{L}_{\mathrm{sft}} + \lambda \mathcal{L}_{temp}.
\end{equation}
Here, $\lambda$ is a fixed coefficient after averaging the auxiliary loss over transition pairs. Continuity is mainly learned by serializing $s_{t-1}$, $I_{t-1}$, $I_t$, Geo-CoT, and $s_t$ as one constrained transition, while Geo-CoT and the 5-frame reset prevent noisy residuals from causing unbounded drift.

\section{Technical Evaluation}
\label{sec:technical_evaluation}

\subsection{Experimental Settings}
\label{sec:experimental_settings}

\paragraph{Dataset and Evaluation Protocol.}
We conduct a systematic evaluation of \model{} on the RoadSceneState dataset, which comprises 2,341 high-resolution video clips. To ensure a robust assessment of generalization capabilities, the dataset is partitioned into training, validation, and test sets following a 15:1:2 ratio. Adhering to the evaluation protocol defined in Table~\ref{tab:main_results}, we report Accuracy (Acc), Precision (P), and Recall (R) across six primary dimensions of structured road understanding: \textit{Lane Count}, \textit{Ego-lane Index}, \textit{Structural Feasibility}, \textit{Vehicle Feasibility}, \textit{Obstacle Feasibility}, and \textit{Global Feasibility}, alongside the aggregated \textit{Overall} performance. These hierarchical metrics provide a comprehensive perspective on the model's ability to perform fine-grained symbolic state estimation.

\begin{table}[t]
\centering
\caption{Efficiency--effectiveness comparison between 4B and 8B Thinking variants under the same state-aware paradigm.}
\label{tab:efficiency_tradeoff}
\small
\renewcommand{\arraystretch}{1.1}
\resizebox{\columnwidth}{!}{%
\begin{tabular}{lcccc}
\toprule
\textbf{Model Variant} & \textbf{VRAM (GB)} & \textbf{Latency (s)} & \textbf{FPS} & \textbf{Mean Acc. (\%)} \\

\midrule
Qwen3-VL-4B-Thinking & 13.0 & 22.6 & 0.088 & 92.10 \\
\model{} (8B) & 21.8 & 23.5 & 0.085 & 94.33 \\
\bottomrule
\end{tabular}
}
\end{table}

\subsection{Main Results}
\label{sec:main_results}

Table~\ref{tab:main_results} presents a comprehensive quantitative evaluation of our proposed \model{} against state-of-the-art baseline Vision-Language Models (VLMs) on the RoadSceneState dataset. 

\paragraph{Benchmark Performance Analysis.}

The experimental results demonstrate that \model{} (backboned by Qwen3-VL-8B-Thinking) achieves state-of-the-art performance with an Overall Accuracy of 94.33\% and a balanced Macro-Precision/Recall of 81.04\% / 81.48\%. Notably, \model{} exhibits exceptional stability in safety-critical dimensions, such as \textit{Obstacle Feasibility} (99.70\% Acc, 99.85\% Recall) and \textit{Structural Feasibility} (94.31\% Acc, 94.25\% Recall). The high recall on sparse obstacle cases shows that Geo-CoT reasoning guides the model to identify rare hazards rather than over-fitting to the majority ``clear'' class, while the structured transition design filters transient occlusions or camera-vibration spikes to avoid short-lived lane-count flips. Furthermore, the Global Feasibility Accuracy of 90.34\% underscores the efficacy of our logic-closure mechanism, ensuring that the final output maintains high consistency with the underlying geometric and temporal constraints.

\begin{table*}[t]
\centering
\caption{Quantitative evaluation of \model{} against state-of-the-art VLMs on the \textbf{RoadSceneState Benchmark}. Performance is measured across six dimensions using Accuracy (Acc), Macro-Precision (P), and Macro-Recall (R); all values are reported as percentages.}
\label{tab:main_results}
\normalsize
\renewcommand{\arraystretch}{1.3}
\setlength{\tabcolsep}{2pt}
\footnotesize
\resizebox{\textwidth}{!}{%
\begin{tabular}{l | ccc | ccc | ccc | ccc | ccc | ccc | ccc}
\toprule
\multirow{2}{*}{\textbf{Model}} & \multicolumn{3}{c|}{\textbf{Lane Count}} & \multicolumn{3}{c|}{\textbf{Ego-lane Index}} & \multicolumn{3}{c|}{\textbf{Struct Feasibility}} & \multicolumn{3}{c|}{\textbf{Vehicle Feasibility}} & \multicolumn{3}{c|}{\textbf{Obstacle Feasibility}} & \multicolumn{3}{c|}{\textbf{Global Feasibility}} & \multicolumn{3}{c}{\textbf{Overall}} \\
\cmidrule(lr){2-4} \cmidrule(lr){5-7} \cmidrule(lr){8-10} \cmidrule(lr){11-13} \cmidrule(lr){14-16} \cmidrule(lr){17-19} \cmidrule(lr){20-22}
& Acc & P & R & Acc & P & R & Acc & P & R & Acc & P & R & Acc & P & R & Acc & P & R & Acc & P & R \\
\midrule
RoadSceneBench & 
- & 63.37 & 65.89 & 
- & 75.44 & 84.67 & 
- & - & - & 
- & - & - & 
- & - & - &  
- & 83.83 & 84.69 & 
- & 74.21 & 78.42 \\
LLaVA-Next-8B & 
76.96 & 65.66 & 49.65 & 
92.66 & 68.23 & 55.99 & 
91.14 & 91.01 & 90.96 & 
90.51 & 64.63 & 65.84 & 
99.79 & 83.25 & 66.65 & 
87.22 & 86.12 & 86.88 & 
89.71 & 76.48 & 69.33 \\
LLaVA-1.5-7B & 
86.33 & 73.10 & 67.09 & 
94.18 & 54.38 & 49.06 & 
91.56 & 91.84 & 91.02 & 
93.08 & 73.20 & 61.69 & 
98.61 & 55.67 & 82.68 & 
88.82 & 87.84 & 88.48 & 
92.10 & 72.67 & 73.34 \\
Qwen2.5-VL-7B & 
52.49 & 52.14 & 44.33 & 
67.26 & 74.07 & 64.75 & 
78.19 & 81.44 & 75.79 & 
94.22 & 90.04 & 60.64 & 
99.83 & 99.92 & 66.67 & 
77.34 & 77.40 & 72.75 & 
78.22 & 79.17 & 64.15 \\
Qwen3-VL-4B-Thinking &
90.20 & 72.85 & 66.94 &
92.50 & 65.71 & 61.83 &
91.80 & 91.42 & 91.35 &
93.50 & 81.76 & 70.92 &
99.40 & 77.31 & 99.18 &
85.20 & 84.06 & 84.73 &
92.10 & 78.85 & 79.16 \\
\midrule

\rowcolor{TabBG}\textbf{\model{}} & 
\textbf{91.90} & \textbf{74.07} & \textbf{67.36} & \textbf{94.60} & \textbf{66.90} & \textbf{63.36} & 
\textbf{94.31} & \textbf{94.05} & \textbf{94.25} & 
\textbf{95.15} & \textbf{84.06} & \textbf{73.87} & 
\textbf{99.70} & \textbf{78.13} & \textbf{99.85} & 
\textbf{90.34} & \textbf{89.00} & \textbf{90.18} & 
\textbf{94.33} & \textbf{81.04} & \textbf{81.48} \\
\bottomrule
\end{tabular}
}
\end{table*}

\subsection{Ablation Study}
\label{sec:ablation}

To systematically investigate the contribution of our proposed modules, we conduct an ablation study with four incremental configurations (M1--M4), as detailed in Table~\ref{tab:ablation}.

\begin{itemize}
    \item \textbf{M1 (Vanilla Perception):} A standard VLM baseline fine-tuned for single-frame road description without structured reasoning.
    \item \textbf{M2 (Explicit Reasoning):} Incorporates the \textbf{Geo-CoT} reasoning chain based on a single image, lacking historical state context.
    \item \textbf{M3 (Direct Transition):} An autoregressive variant that predicts the current state $s_t$ given $\{I_{t-1}, I_t, s_{t-1}\}$ but bypasses the intermediate Geo-CoT logic.
    \item \textbf{M4 (\model{} Full):} Our complete paradigm featuring both the autoregressive state transition and the DAG-guided Geo-CoT distillation.
\end{itemize}

\paragraph{Impact of Temporal and Historical Context.} Comparing M2 and M4 demonstrates the benefit of incorporating temporal and historical context into structured road-state estimation. 
M2 performs Geo-CoT reasoning from a single frame, whereas M4 additionally incorporates the previous frame $I_{t-1}$ and the previous road state $s_{t-1}$ to update the current state $s_t$. 
Together, these inputs increase mean accuracy from 81.83\% to 94.33\% (+12.50 percentage points), with gains across all six state dimensions.

\paragraph{Significance of Geo-CoT Scaffolding.} The performance gap between M3 and M4 (from 88.71\% to 94.33\% Mean Acc) validates the necessity of Geo-CoT as a logical scaffold. While M3 attempts to map sequential images directly to a JSON state, it often produces hallucinated transitions that violate physical constraints. In contrast, the Geo-CoT in M4 enforces a deterministic chain of evidence, from geometric grounding to logic closure, significantly boosting the accuracy of complex tasks like \textit{Lane Count} calculation (+12.58 percentage points).

\begin{table*}[t]
\centering
\caption{Ablation Study on Geo-CoT Reasoning and Historical State Context. Metrics report the \textbf{Accuracy (Acc)} for each state dimension, with all accuracy values reported as percentages. M1 does not output the three feasibility dimensions; therefore, its aggregate mean is omitted and should not be compared directly with M2--M4. M4 represents our full proposed \model{} framework.}
\label{tab:ablation}
\normalsize
\renewcommand{\arraystretch}{1.3} 
\resizebox{\textwidth}{!}{
\begin{tabular}{lcc | cccccc | c}
\toprule
\multirow{2}{*}{\textbf{Variant}} & \multicolumn{2}{c|}{\textbf{Components}} & \multicolumn{6}{c|}{\textbf{State Dimension Accuracy (\%)}} & \multirow{2}{*}{\textbf{Mean Acc. (\%)}} \\
\cmidrule(lr){2-3} \cmidrule(lr){4-9}
& Geo-CoT & Historical State & Lane Count & Ego Index & Str. Feas. & Veh. Feas. & Obs. Feas. & Global Feas. & \\
\midrule
M1 (Single-Frame Description) & $\times$ & Single-frame & 
68.69 & 71.56 & - & - & - & 61.94 & -- \\
M2 (Single-Frame + CoT) & $\checkmark$ & Single-frame &
68.10 & 75.36 & 79.33 & 91.31 & 99.24 & 77.64 & 81.83 \\
M3 (State Prediction (Direct)) & $\times$ & Multi-frame & 
79.32 & 82.03 & 88.48 & 96.12 & 99.75 & 86.54 & 88.71 \\
\midrule
\rowcolor{TabBG}\textbf{M4 (\model{})} & $\checkmark$ & \textbf{Multi-frame} & \textbf{91.90} & \textbf{94.60} & \textbf{94.31} & \textbf{95.15} & \textbf{99.70} & \textbf{90.34} & \textbf{94.33} \\
\bottomrule
\end{tabular}
}
\end{table*}

\paragraph{Backbone versus structured reasoning.}
To disentangle the gain from the Thinking-capable backbone and the gain from Geo-CoT-style structured reasoning, we further compare four Qwen3-VL-8B variants under the same state-aware evaluation protocol in Table~\ref{tab:thinking_variant}. The Instruct backbone benefits substantially from Geo-CoT supervision, improving Mean Acc from 86.50\% to 92.95\%. The Thinking backbone without explicit Geo-CoT reaches 88.71\%, showing that stronger native reasoning alone is helpful but insufficient for topology-sensitive map production. Our full \model{} achieves 94.33\% Mean Acc, improving by 5.62 percentage points over the state-aware Thinking variant without Geo-CoT. This indicates that the main practical benefit comes from Thinking-based structured reasoning internalized through Geo-CoT, especially for lane topology and feasibility dimensions where temporal consistency and logical closure are most critical.

\begin{table*}[t]
\centering
\caption{Model-variant comparison of Thinking capability and Geo-CoT supervision. All accuracy values are reported as percentages.}
\label{tab:thinking_variant}
\normalsize
\renewcommand{\arraystretch}{1.2}
\resizebox{\textwidth}{!}{
\begin{tabular}{lccccccc}
\toprule
\textbf{Model Variant} & \textbf{Lane Count} & \textbf{Ego Index} & \textbf{Str. Feas.} & \textbf{Veh. Feas.} & \textbf{Obs. Feas.} & \textbf{Global Feas.} & \textbf{Mean Acc. (\%)} \\
\midrule
Qwen3-VL-Instruct-8B (No CoT) & 77.50 & 81.10 & 83.20 & 94.00 & 99.00 & 84.20 & 86.50 \\
Qwen3-VL-Instruct-8B (Geo-CoT) & 92.32 & 94.35 & 84.89 & 95.57 & 99.54 & 91.01 & 92.95 \\
Qwen3-VL-Thinking-8B (No CoT) & 79.32 & 82.03 & 88.48 & 96.12 & 99.75 & 86.54 & 88.71 \\
\midrule
\rowcolor{TabBG}\textbf{\model{}} & \textbf{91.90} & \textbf{94.60} & \textbf{94.31} & \textbf{95.15} & \textbf{99.70} & \textbf{90.34} & \textbf{94.33} \\
\bottomrule
\end{tabular}
}
\end{table*}

\paragraph{Robustness and Mechanism Ablation.}
Table~\ref{tab:mechanism_robustness} reports the Logical Override ablation together with the single-frame M2 reference. Removing Logical Override lowers Mean Acc from 94.33\% to 91.82\%. The M2 row is repeated from Table~\ref{tab:ablation} to show the effect of removing temporal input; it is not a test of the five-frame reset. Measuring the reset interval requires a separate long-sequence experiment.

\begin{table*}[t]
\centering
\caption{Accuracy (\%) for the Logical Override ablation and the single-frame M2 reference.}
\label{tab:mechanism_robustness}
\small
\renewcommand{\arraystretch}{1.1}
\resizebox{\textwidth}{!}{%
\begin{tabular}{lccccccc}
\toprule
\textbf{Variant} & \textbf{Lane Count} & \textbf{Ego Index} & \textbf{Str. Feas.} & \textbf{Veh. Feas.} & \textbf{Obs. Feas.} & \textbf{Global Feas.} & \textbf{Mean Acc.} \\
\midrule
w/o Logical Override & 88.12 & 90.30 & 91.05 & 92.50 & 99.12 & 89.80 & 91.82 \\
w/o Historical State Feedback (M2) & 68.10 & 75.36 & 79.33 & 91.31 & 99.24 & 77.64 & 81.83 \\
\rowcolor{TabBG}\textbf{\model{} Full} & \textbf{91.90} & \textbf{94.60} & \textbf{94.31} & \textbf{95.15} & \textbf{99.70} & \textbf{90.34} & \textbf{94.33} \\
\bottomrule
\end{tabular}%
}
\end{table*}

\section{Production Deployment and Application Impact}
\label{sec:production_deployment}

\system{} has been integrated into the map-updating pipeline of Baidu Maps and supports routine lane-level updates across more than 360 cities. In this workflow, geometry-aware estimates of lane count and ego-lane position provide structured scene cues for downstream map-change monitoring and observation selection. In particular, these estimates help prioritize observations with favorable road coverage and reduce ambiguity in lane-level map updates. Under production stress testing, the pipeline achieves an inference completion rate above $99\%$, indicating that the system can operate reliably at the scale required by the offline updating workflow.

\paragraph{Cascaded Production Inference.}
\label{sec:cascade}
Applying an 8B reasoning model to every incoming frame would be unnecessarily expensive at daily-update scale (R4 in Section~\ref{sec:application_context}). In practice, the difficulty distribution of crowd-sourced observations is highly imbalanced: most frames contain clearly visible road markings and unambiguous lane configurations, whereas only a smaller subset involves occlusion, worn markings, ramps, temporary obstacles, or other cases that benefit from state-aware reasoning. We therefore adopt a cascaded inference strategy rather than applying the same model uniformly. Approximately $85\%$ of frames are handled by lightweight models, while the remaining $\sim15\%$ are routed to \model{}. Both branches produce the same structured road-state representation, so the routing policy is decoupled from downstream validation, auditing, and map-fusion modules and can be adjusted independently as production requirements evolve.

We further compare the deployed 8B \model{} with a Qwen3-VL-4B-Thinking variant under the same state-aware inference paradigm. As shown in Table~\ref{tab:efficiency_tradeoff}, reducing the backbone from 8B to 4B decreases end-to-end latency only from 23.55 s to 22.6 s per two-frame state transition, while the equivalent input-frame throughput increases from 0.085 to 0.088 FPS. In contrast, mean accuracy decreases from $94.33\%$ to $92.10\%$. The degradation is more pronounced in challenging scenarios, where the 4B model is consistently 4--7 percentage points below the 8B model, including an approximately 5-point drop in lane-topology accuracy under occlusion. These results show that reducing model size provides only limited serving gains in our pipeline while introducing a measurable loss in reliability. We therefore retain the 8B model for the escalated branch and use cascaded routing to control its overall invocation frequency.

\paragraph{Human-in-the-Loop Quality Control.}
Each predicted state is first checked against the dependency constraints used by Geo-CoT. Samples are sent to human review when the prediction violates a constraint, has low model confidence, or exhibits a topology not covered by the current constraint set. This design separates automated inference from the final acceptance decision: the model improves processing efficiency, while uncertain or out-of-distribution cases remain subject to explicit verification.

Structured output is important for keeping this review process efficient (R3 in Section~\ref{sec:application_context}). Instead of inspecting free-form scene descriptions, auditors are presented with a compact road state consisting of lane count, ego-lane index, and the three lane-change feasibility factors, together with the corresponding Geo-CoT trace. Because feasibility is decomposed into structural, vehicle, and obstacle factors, an auditor can localize the source of a negative decision and revise the relevant component without re-annotating the entire state. Audited corrections are subsequently accumulated as additional supervision for model improvement, forming a practical feedback loop in which difficult production cases can be incorporated into later training iterations.

\paragraph{Deployment Efficiency.}
We profile the escalated branch on a single NVIDIA A800 GPU using bf16 precision. The Qwen3-VL-8B backbone occupies 21.83 GB of VRAM. For a two-frame sample corresponding to one state transition, the average end-to-end latency is 23.55 s, consisting of 0.83 s for image encoding, 18.20 s for autoregressive Geo-CoT generation, and 4.52 s for final structured-state generation. Geo-CoT produces approximately 485 tokens per transition on average. The autoregressive reasoning stage therefore accounts for the majority of inference time. 
In the asynchronous, offline map-updating workflow, cascaded routing and parallel serving enable the deployed system to meet the required production throughput.

\paragraph{Production Impact.}
With structured road states and reasoning traces, operator throughput increases from 800 to 1{,}150 images per person-day, corresponding to a $43.8\%$ increase in review throughput, while the measured labor cost per image decreases by $30.4\%$. Together with the reduced frequency of invoking the full VLM under cascaded inference, these improvements correspond to an approximately $30\%$ reduction in the overall production cost of this stage of the map-updating workflow. These results indicate that the main value of \system{} in production lies in combining selective model inference with structured human verification, enabling higher processing efficiency while retaining explicit quality control for difficult cases.

\paragraph{Operational Transferability.}
The deployment experience also suggests that the value of RoadState-VLM comes from its position within the update workflow rather than from isolated recognition accuracy. The shared JSON schema allows lightweight models, the escalated VLM, the constraint checker, and the audit console to exchange the same state object, so model replacement does not require changes to map-change fusion. This separation also provides a controlled fallback: when confidence is low or a topology falls outside the covered rules, the system can preserve the observation for manual review instead of committing an unsupported map change. 

\section{Lessons Learned from Development and Deployment}
\label{sec:lessons}

\paragraph{Explicit domain constraints and temporal context fix structural errors more effectively than increasing model size.}
Our two main failure modes in early trials---physically invalid lane topologies and frame-to-frame prediction flicker---could not be resolved simply by scaling up the VLM. Increasing the backbone from 4B to 8B improves mean accuracy by 2.23 percentage points (Table~\ref{tab:efficiency_tradeoff}), whereas adding temporal state context (+12.50 points, M2 to M4) and DAG-guided Geo-CoT supervision (+5.62 points, M3 to M4) yields much larger gains on these errors (Section~\ref{sec:ablation}). A larger 8B backbone still helps interpret heavily occluded scenes, but encoding domain rules and temporal continuity provides higher reliability gains without increasing per-frame serving cost.

\paragraph{Cascaded routing makes slow reasoning VLMs practical by reserving them for hard cases.}
Running an 8B reasoning model (23.55~s per transition) on every incoming frame is too costly at daily-update volume. Because roughly $85\%$ of crowdsourced frames have clear markings that lightweight classifiers handle reliably, routing only the remaining $\sim$$15\%$ of occluded or ambiguous cases to \model{} retains its accuracy advantage on difficult scenes while keeping average compute per frame low (Section~\ref{sec:cascade}). In practice, refining the triage criteria for escalation proved just as impactful for overall cost and quality as improving the VLM itself.

\paragraph{Autoregressive state tracking requires periodic resets and human fallback to prevent error drift.}
Conditioning each prediction on the previous state $\hat{s}_{t-1}$ suppresses single-frame noise, but an early misprediction can otherwise carry over to subsequent frames. To bound error propagation, \system{} resets autoregressive inference every five frames by re-estimating the state directly from the current frame. In addition, predictions that violate domain constraints, exhibit low confidence, or involve rare topologies outside our rule set are automatically routed to human review.

\paragraph{A shared structured state schema simplifies both pipeline integration and human review.}
Instead of passing free-form text, every stage in \system{}---lightweight classifiers, \model{}, the deterministic constraint checker, the audit console, and map-change fusion---exchanges the same structured JSON schema. Using one shared representation allows individual models to be upgraded or replaced without modifying downstream modules. Moreover, decomposing lane-change feasibility into separate structural, vehicle, and obstacle fields enables human auditors to locate and fix a single erroneous factor rather than relabeling the entire state.

\section{Conclusion}
\label{sec:concl}
In this paper, we present \system{}, a vision-language system deployed in Baidu Maps' map-updating workflow to generate structured lane-level semantics.
Its core model, \model{}, casts road-scene understanding as autoregressive structured state estimation. It conditions each prediction on the previous symbolic state to improve temporal stability and uses DAG-guided Geo-CoT supervision to encode dependencies among road-state variables, improving structural validity and interpretability in lane-level road-state prediction.
Experimental results show that \model{} outperforms strong VLM baselines in lane topology and feasibility tasks. 
The system has been integrated into a large-scale commercial mapping platform, supporting routine updates across 360+ cities and achieving a 30\% reduction in overall production cost.
By combining domain-guided reasoning with a structured state interface, selective VLM inference, and human review, \system{} makes VLM-based road-scene understanding practical for industrial-grade lane-level map updating.

\section{Acknowledgments}
This work was supported by Beijing Natural Science Foundation (L231008, L243008), National Key R\&D Program of China (2024YFB2505802), National Natural Science Foundation of China (52472449, 52402499), Independent Research Project of the State Key Laboratory of Intelligent Green Vehicle and Mobility, Tsinghua University (No. ZZ-PY-20250408), the Tsinghua University-Toyota Joint Center, and Tsinghua University--SAIC GM Wuling Joint Research Center.

\bibliography{main}

\appendix
\renewcommand{\topfraction}{0.95}
\renewcommand{\dbltopfraction}{0.95}
\renewcommand{\dblfloatpagefraction}{0.80}
\renewcommand{\textfraction}{0.05}
\setcounter{dbltopnumber}{2}
\twocolumn[{%
  \begin{center}
    {\Large\bf Appendix\par}
  \end{center}
  \vspace{0.8em}
  \centering
  \includegraphics[width=0.92\textwidth]{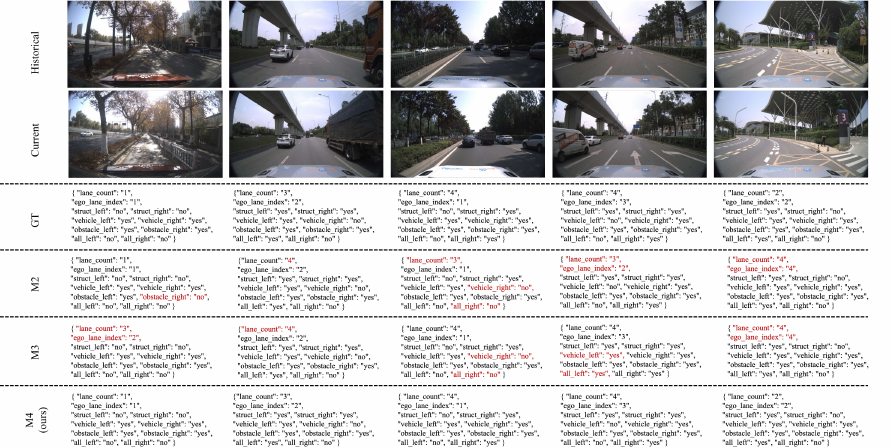}
  \captionof{figure}{Qualitative Case Studies. Representative visualization cases highlighting the complexity of real-world driving scenarios. Red text denotes prediction errors, underscoring the superior reasoning capabilities of our method.}
  \label{fig:case_study}
  \vspace{1.2em}
}]

\section{Training and Inference Implementation Details}
\label{sec:appendix_implementation}

\paragraph{Parameter-Efficient Adaptation.}
We fine-tune Qwen3-VL-8B-Thinking with LoRA to internalize road-specific constraints efficiently. Trainable low-rank matrices with rank $r=8$ and scaling $\alpha=16$ are injected into all LLM linear layers. We set the multimodal input limit to $1,568,000$ pixels and use \texttt{bf16} mixed-precision training.

\paragraph{Baselines and Implementation Details.}
We benchmark \model{} against a suite of representative state-of-the-art open-source VLMs, including LLaVA \cite{liu2024improved,li2024llavanext-strong} and the Qwen series \cite{bai2025qwen25vltechnicalreport,bai2025qwen3}. To reduce discrepancies from task-specific adaptation, all baseline models underwent the same Supervised Fine-Tuning (SFT) protocol on the RoadSceneState dataset. All experiments were implemented within the LLaMA-Factory framework and executed on NVIDIA A800 GPUs. For inference, we adopted deterministic decoding with temperature fixed at $0.0$ and $top\text{-}p$ set to $1.0$.

\section{Qualitative Case Studies of Geo-CoT Reasoning}
\label{sec:appendix_reasoning}

This section presents representative qualitative examples of the proposed Geo-CoT reasoning process. These cases illustrate how the model integrates global geometry, temporal context, and structural constraints to produce logically consistent lane-level conclusions.

\subsection{Comparative Case Studies}
\label{sec:qualitative}

We present representative cases in Figure~\ref{fig:case_study} to demonstrate the distinct advantages of \model{} in handling safety-critical edge cases:

\begin{itemize}
    \item \textbf{Cross-Attribute Logic Closure:} In scenarios where visual cues are conflicting (e.g., a dashed lane line suggests structural feasibility while an adjacent vehicle poses a collision risk), baseline models may output an erroneous ``yes'' based on surface-level features. In contrast, \model{} applies the conjunction $\mathcal{F}_{str} \wedge \mathcal{F}_{veh} \wedge \mathcal{F}_{obs}$, so a vehicle-related violation can correctly make the global feasibility negative.
    \item \textbf{Visual Perturbation Resilience:} Under temporary visual degradation, such as occlusion by a large vehicle or glare-induced blur, the previous state provides a temporal anchor. In scenarios without merges, splits, or lane changes, this continuity helps preserve lane-count and ego-position estimates instead of allowing unsupported frame-to-frame changes.
\end{itemize}

\subsection{Geo-CoT Reasoning Traces}

\begin{figure*}[t]
  \centering
  \includegraphics[width=0.92\linewidth]{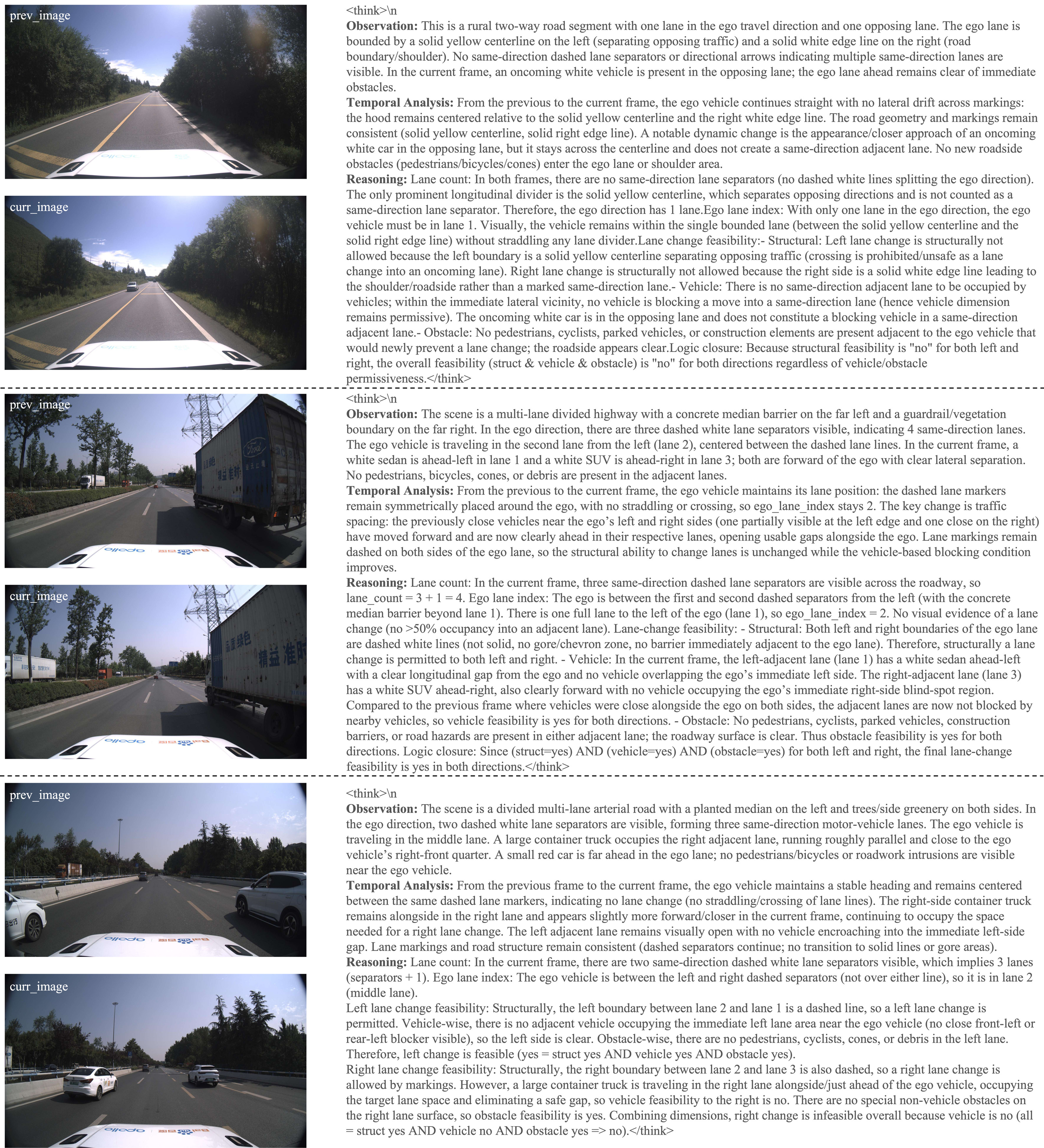}
  \caption{\textbf{Qualitative Case Studies of Geo-CoT Reasoning.} Representative samples illustrating the hierarchical reasoning process of \textsf{\model{}} across diverse driving environments. Each case demonstrates the model's logic through \textbf{Global Observation} of static road geometry, \textbf{Temporal Analysis} of ego-motion, and \textbf{Logical Deduction} enforcing axiomatic constraints.}
  \label{fig:cot-case-appendix}
\end{figure*}

Across diverse driving environments, the cases in Figure~\ref{fig:cot-case-appendix} demonstrate the interpretability and stability of the hierarchical reasoning pipeline. Additional qualitative results and runnable examples are available at: {\color{blue}\url{https://github.com/eadst/IAAI-2027-RoadState-VLM}}

\end{document}